%% file: main.tex
\documentclass[journal]{IEEEtran}
\usepackage{ifpdf}
\usepackage{cite}
\usepackage{algorithmic}
\usepackage{array}
\usepackage{amsmath}
\usepackage{amssymb}
\usepackage{booktabs}
\usepackage{multirow}
\usepackage{dblfloatfix}
\usepackage{url}
\usepackage{float}
\usepackage[table,xcdraw]{xcolor}
\ifCLASSINFOpdf
  \usepackage[pdftex]{graphicx}
\else
  \usepackage[dvips]{graphicx}
\fi

\ifCLASSOPTIONcompsoc
 \usepackage[caption=false,font=normalsize,labelfont=sf,textfont=sf]{subfig}
\else
 \usepackage[caption=false,font=footnotesize]{subfig}
\fi

\ifCLASSOPTIONcaptionsoff
 \usepackage[nomarkers]{endfloat}
\let\MYoriglatexcaption\caption
\renewcommand{\caption}[2][\relax]{\MYoriglatexcaption[#2]{#2}}
\fi

\begin{document}
\title{Segment Any Class (SAC): Multi-Class Few-Shot Semantic Segmentation via Class Region Proposals}

\author{Hussni Mohd Zakir, Eric Tatt Wei Ho \\
Department of Electrical \& Electronics Engineering\\
Universiti Teknologi PETRONAS \\
32610 Seri Iskandar, Perak, Malaysia
}

% need change
% \markboth{Journal of \LaTeX\ Class Files,~Vol.~14, No.~8, August~2015}%
% {Shell \MakeLowercase{\textit{et al.}}: Bare Demo of IEEEtran.cls for IEEE Journals}

\maketitle

\begin{abstract}
The Segment-Anything Model (SAM) is a vision foundation model for segmentation with a prompt-driven framework. 
SAM generates class-agnostic masks based on user-specified instance-referring prompts. However, adapting SAM for automated segmentation —where manual input is absent— of specific object classes often requires additional model training. We present Segment Any Class (SAC), a novel, training-free approach that task-adapts SAM for Multi-class segmentation. SAC generates Class-Region Proposals (CRP) on query images which allows us to automatically generate class-aware prompts on probable locations of class instances. CRPs are derived from elementary intra-class and inter-class feature distinctions without any additional training. Our method is versatile, accommodating any N-way K-shot configurations for the multi-class few-shot semantic segmentation (FSS) task. Unlike gradient-learning adaptation of generalist models which risk the loss of generalization and potentially suffer from catastrophic forgetting, SAC solely utilizes automated prompting and achieves superior results over state-of-the-art methods on the COCO-20i benchmark, particularly excelling in high N-way class scenarios. SAC is an interesting demonstration of a prompt-only approach to adapting foundation models for novel tasks with small, limited datasets without any modifications to the foundation model itself. This method offers interesting benefits such as intrinsic immunity to concept or feature loss and rapid, online task adaptation of foundation models.
\end{abstract}

\IEEEpeerreviewmaketitle

\section{Introduction}
\

Generalizable deep neural network models are key to proliferating artificial intelligence (AI) in myriad applications. A generalizable AI model is trained once and rapidly deployed for a variety of different use cases or objectives. The asymmetry in development versus deployment effort vastly improves the efficiency of data use, amortizes the cost-effort and energy consumption of model training over a host of different use cases, and thus encourages investment in high-quality feature learning processes. Foundation models are large and deep generative neural network models trained on extensive and comprehensive datasets. Various foundation models have recently been developed for different data domains such as vision, language, and audio. Foundation models have demonstrated promising potential as generalizable AI models that are adaptable to different tasks or data domains. While the models are typically trained on several specific objectives, they have also demonstrably excelled on novel tasks. For example, large language models (LLM) trained for sequence prediction can be prompted to solve question-answering tasks or language translation queries and large diffusion models may generate images with previously unseen combinations of objects and poses. 
 
Foundation models appear to be good, generalizable feature models. They demonstrate that deep neural networks can learn complex, correct, and rich semantic representations of knowledge when training datasets exceed a certain scale. Model features are learned through a diverse training regime typically comprising an initial unsupervised training of a generative model, followed by supervised learning on multiple different tasks and further fine-tuning through a combination of self-consistency learning and reinforcement learning with human feedback. Foundation models are also generative models and therefore generate outputs in response to inputs that may be interpreted as random seeds, latent vectors that embed semantic concepts, or cross-domain signals that serve as conditional vectors. The generalizability of learned features may be improved by scaling up the size and diversity of data and supervised and self-consistency learning processes but the topic of how to tune foundation models to a novel task remains of continued research interest, especially for use cases involving small labeled training datasets or concept drift. The key challenge when adapting foundation models to novel tasks lies in achieving the right balance when integrating new task-specific features and semantics into the knowledge and feature representation of the foundation models in a way that synergizes but does not displace previously learned concepts. The supervised learning paradigm appears ill-aligned to achieve this goal; direct updates of model weights do not provide a framework that prevents newly learned features and semantics from overwriting older ones nor can it guide toward a revised representation that better captures the feature and semantic similarities and differences of the old and new. Starting supervised learning from scratch for new datasets is also not desired as it misses out on the rich information learned from the extensive training of foundational models.

Nevertheless, gradient learning has long been considered a cornerstone for adapting deep learning models to new tasks or new datasets as evident from early strategies like transfer learning to contemporary techniques such as delta-tuning with low-rank adaptations and adapters for large foundation models. Adaptation of pre-trained networks allows beneficial reuse of a huge corpus of data, not necessarily related to the task or dataset of interest, to repurpose good feature representations while focally applying the smaller, task-related dataset towards learning correct decisions for the new task. This logic supports the development of generalizable large foundation models and ultimately aims towards enhancing the accuracy of new tasks on small and restricted datasets. However, the optimal way to adapt or tune foundation models to new tasks with limited datasets remains unknown. Gradient-learning approaches suffer from the possibility of catastrophic forgetting as tuning network weights on new datasets or new tasks also risks displacing previously trained semantics. 

Arising from generative neural network models, prompts provide a new, alternative approach to adapting neural networks to new tasks. To gain intrinsic immunity to catastrophic forgetting, we argue against modifying the network weights of foundation models. In cases where intentional prompt curation suggests that the foundational model can accomplish a specialized task, we suggest that prompts play the role of providing the organizational principles of the features that enable foundation models to perform a new task or adapt to a new dataset. Therefore, the semantic and feature relationships present in new datasets must somehow be incorporated into the prompt input and not through modification of the foundation model weights by gradient learning. 

In the present study, we leverage DINOv2 and SAM, trained on the extensive LVD-142M and SA-1B datasets, respectively with our method to achieve state-of-the-art results on Multi-class FSS on COCO-$20^i$. Despite not applying any gradient-learning, our results show state-of-art performance in multiclass few shot segmentation and we argue for automated prompt generation as a viable alternative to gradient-based network adaptation.

% If foundation models already contain generalizable features that can be reorganized for any task, only the prompt inputs should be modified.  

% We propose that since foundation models inherently possess rich semantics, altering their weights through gradient-based learning may be suboptimal for achieving optimal results.  In cases where intentional prompt curation suggests that the foundational model can accomplish a specialized task, employing a prompt engineering approach proves more effective than gradient-based methods. Here, we demonstrate the strength of this concept by overcoming the multiclass segmentation task against fully-supervised few-shot segmentation models.

% \textcolor{red}{In this paper, To further illustrate this paradigm, we take the example of Large Language Models (LLM) the prompt should contain information about how to assemble words for a task or the grammar of a new language in translation whereas for Large Image Models the prompt should contain rules governing the Markov random fields of specific objects, textures or scenes.Here, we demonstrate the strength of this concept by overcoming the multiclass segmentation which outperforms few shot trained models. Can revert to previous dataset training.
% Evidence on the argument of to prompt or to learn?
% Fundamental question of representation or retrieval & organization?}

\section{Related Work}

%\subsection{Promptable Foundation Models}

Foundation models forged through extensive training on vast datasets, have become renowned for their remarkable generalization abilities and suitability for a variety of downstream tasks. In natural language processing (NLP), models like BERT \cite{Devlin2018BERT:Understanding}, GPT \cite{Brown2020LanguageLearners}, and LLAMA \cite{Touvron2023LLaMA:Models} exemplify this capability by adeptly transferring knowledge to new tasks through domain-specific prompts, even when these tasks diverge from their original training objectives. In the realm of computer vision, innovations like Masked Autoencoders (MAE) \cite{He2021MaskedLearners} draw inspiration from masked language modeling, employing transformers to generate masked image patches. On the other hand, DINOv2 \cite{Oquab2023DINOv2:Supervision} refines feature learning through discriminative self-supervised approaches, broadening the applicability of learned features.

A common trend among recent foundation models is the integration of a human interface via prompting. Prompts serve as intuitive input guides that shape the model's output features or content. The concept of prompting can be traced back to Generative Adversarial Networks (GANs) \cite{Goodfellow2014GenerativeNetworks}, where inputs to the generator diversify outputs through mechanisms like random seeds, latent vectors representing semantic concepts, or conditional signals \cite{Mirza2014ConditionalNets}. In the domain of language, LLMs interpret natural language prompts, formatted as questions or commands, to perform new tasks. Similarly, multi-modal models like Stable Diffusion \cite{Rombach2021High-ResolutionModels} process text prompts to generate images reflective of the described content. 

%Sequence prompt optimization, encompassing both text and audio, has been demonstrated to enhance the quality and accuracy of outputs from foundational models \cite{Scao2021HowWorth}. Notable techniques such as Chain-of-Thought prompting \cite{Wei2022Chain-of-ThoughtModels} elicit intermediate reasoning steps from models, thereby refining the coherence and quality of generated outputs. Methods like Hypothesis Testing Prompting \cite{Li2024WhatHypothesis} further instruct models to generate outputs following explicit procedural guidelines.

Beyond sequence-based prompts, foundation models in vision are also receptive to prompts in the image domain, such as reference images or specified points and boxes. Image-domain prompts impose new task constraints by delineating reference regions that guide model behavior on query images. The Segment-Anything Model (SAM) \cite{Kirillov2023SegmentAnything}, trained on over 1 billion masks and 11 million images, excels in zero-shot and class-agnostic segmentation through such inputs. Expanding on this paradigm, SEEM \cite{Zou2023SegmentOnce} integrates audio and text prompts as part of its multi-modal approach. And most recently, Segment-Anything Model 2 \cite{Ravi2024SAMVideos} was developed which improved upon SAM by extending promptable segmentation to videos. It has a transformer architecture with streaming memory which allows for real-time video processing. Nevertheless, unlike LLMs, which typically excel across diverse language tasks, vision foundation models often struggle with atypical images that deviate from natural scenes \cite{Chen2023SAMMore}, necessitating fine-tuning for optimal performance on novel objectives.

To enhance model generalizability for machine vision tasks, models are developed for in-context learning, where a set of input-output examples are presented to the model together with the target image. The model extracts the object features from the input-output support set and uses it to perform the task on the target image. For instance, after providing several labeled cat images, a model might be tasked with segmenting a new cat in a target image. In-context learning models like DINOv \cite{Li2023VisualSegmentation} are adept at processing both specific and generic prompts to isolate objects of interest or generalize to semantically similar objects in new contexts. VRP-SAM \cite{Sun2024VRP-SAM:Prompt} introduces an innovative trainable prompt-encoding module that derives embeddings from visual reference image-label pairs. Solutions like PerSAM \cite{Zhang2023PersonalizeShot} tailor SAM for one-shot segmentation, while Matcher \cite{Liu2023Matcher:Matching} pioneers frameworks for adapting foundational models for single-class few-shot semantic segmentation (FSS). Other works, such as \cite{Sun2023ExploringLearning} and \cite{Zhang2023WhatLearning}, delve into factors that influence in-context learning performance, emphasizing the critical role of support image selection. 

Shifting from single- to multi-class FSS introduces a new set of challenges. Multi-class FSS models show marked decline in segmentation accuracy as the number of classes grows. This performance decline is largely attributed to the increased uncertainty regarding the presence of various class objects within an image. In single-class FSS evaluations, the target class is always present in the query image; conversely, multi-class evaluations may feature query images absent of certain classes, leading to a heightened susceptibility to false positives and a more significant impact on metric performance in these settings. State-of-art models multi-class FSS models such as Dense pixel-wise Cross-query-and-support Attention weighted Mask Aggregation (DCAMA) \cite{Shi2022DenseSegmentation} and Label-Anything (LA) \cite{DeMarinis2024LabelPrompts} attempt these challenges with specialized mask decoding and prompt encoding models that leverage on pre-trained or foundation image feature models. To support this trend, there are benchmarks that compare the adaptability of various foundation models for the FSS task \cite{BensaidAModels}.

\section{Model}
\input{figures/sac}
Our proposed SAC is illustrated in Fig\ref{fig:sac}, and operates in two main phases: \textbf{Support Feature Extraction} and \textbf{Mask Prediction with Class Region Proposals}. Below, we describe each phase in detail.

\subsection{Support Feature Extraction}
Given a set of support images $\mathcal{I}=\{I_1,...,I_N\}\ \in \mathbb{R}^{N \times H \times W \times 3}$, each annotated with semantic labels $\mathcal{M}=\{M_1,...,M_N\}\ \in \mathbb{R}^{N \times H \times W \times C} $, the first phase aims to construct Class-Representative Feature Arrays (CRFA) that capture the essential class-specific characteristics.

Each support image is passed through the DINOv2, $\text{Enc}_I$, model to obtain its image embedding as seen in Equation (\ref{eq:enc}), which serves as a high-dimensional representation of the image in a feature space. DINOv2, a self-supervised vision model, is used here for its ability to learn meaningful image representations without relying on labeled data. 
\begin{equation}
  \mathcal{F}=\text{Enc}_I(\mathcal{I})\in \mathbb{R}^{N \times h \times w \times embed\_dim}
  \label{eq:enc}
\end{equation}
Next, The embeddings $\mathcal{F}$ are then stratified according to their associated class labels, effectively grouping the embeddings of images belonging to the same class. Equation (\ref{eq:feat_r}) shows this operation where $\circ$ is spatial-wise multiplication. This stratification ensures that the features corresponding to each class are separated for subsequent processing.
\begin{equation}
  \mathcal{R}^c = \{R^c_i\}^N_{i=1} = F^c_i \circ M^c_i
  \label{eq:feat_r}
\end{equation}
For each class, we apply clustering techniques (e.g., k-means) to group the embeddings into class-specific feature arrays, $\mathcal{C}^c$, called Class-Representative Feature Arrays (CRFA). These CRFAs represent the key features that define each class in the embedding space. Additionally, we create CRFAs for the background class, using data from unlabeled regions, to make full use of the information available in the support images. The inclusion of background CRFAs is instrumental in reducing false positive rates.

\begin{equation}
  \mathcal{C}^c = \text{cluster}(\text{concat}(\mathcal{R}^c),n\_c) \in \mathbb{R}^{n\_c \times embed\_dim}
  \label{eq:cluster}
\end{equation}
If the pool of support images becomes too large, we apply a memory-efficient clustering approach. Specifically, we perform clustering at the image level before appending a new image's features to the existing class feature arrays. This reduces the size of the feature arrays, minimizing memory usage, and ensures that we do not overwhelm the system with an excessive number of features before clustering again.

\subsection{Mask Prediction with Class Region Proposals}
During inference, given a query image, $I_q$, we move to the second phase, which involves generating Class Region Proposals (CRP) and predicting class masks.

The query image is passed through the DINOv2 model to obtain its embedding $F_q$, analogous to the process applied to the support images. This embedding represents the query image in the same feature space as the CRFAs.

We compute similarity maps $\{S^c_i\}$ by calculating the cosine similarity between the query image embedding and each class's CRFAs. This allows us to measure how similar each pixel in the query image is to the class representative features. The similarity is computed at the pixel level, generating a per-pixel similarity score for each class.
\begin{equation}
  \{S^c_i\}_{i=1}^{n\_cluster}=F_q C^{cT}, S^c_i \in \mathbb{R}^{H \times W}
  \label{eq:sim_map}
\end{equation}
For each class, we assign each pixel in the query image to the class that has the highest cosine similarity score in the similarity map. However, this approach could potentially assign pixels with low intra-class similarity but high inter-class similarity, which may result in inaccurate class predictions. To address this, we perform \textbf{intra-class filtering}:
We use Otsu-thresholding on the cosine similarity values of each class to filter out pixels where the similarity within the class is low. This ensures that only pixels with high similarity to the class prototype are retained in the CRP, thus improving the precision of the class prediction.

In addition to the class-specific region proposals, we generate a Background Region Proposal (BGRP). A class' BGRP captures the regions of the query image that do not belong to any class or belong to other classes. This helps to distinguish between foreground (class-specific) and background (non-class) regions in the final mask predictions.

Using the CRP and BGRP, we automatically generate class-based prompt sets. First, we create a grid of positive and negative point prompts. The positive point prompts correspond to pixels within the CRP, while the negative point prompts are selected from the BGRP. Each positive point is paired with a random negative point to create multiple prompt sets of positive-negative pairs for each class. Furthermore, inspired by \cite{Liu2023Matcher:Matching}, we cluster our positive points based on their spatial locations, thereby creating a set of prompts consisting of cluster centers and a set of prompts of positive points groups based on the cluster centers.

Each class’s prompt sets generate binary masks for the predicted class. The masks' class corresponds to the positive point of the prompt. To ensure that no pixel is assigned to multiple classes (i.e., avoiding multi-class predictions for the same pixel), we apply \textbf{inter-class filtering}. We remove masks that overlap with other class regions in the CRP if the overlap exceeds a certain threshold. Additionally, masks whose average cosine similarity inside them is below a predefined threshold are discarded, ensuring that only high-confidence predictions remain.

\section{Experiments}

Our experiments were conducted using the Segment-Anything Model (SAM) with the ViT-H backbone and DINOv2 with the ViT-L backbone. For input resolution, we used $518\times518$ for DINOv2 and $1024\times1024$  for SAM, applying rescaling and padding to the images as necessary. We set the parameter $n\_cluster$ to 5. Importantly, there was no additional training performed on the networks in any of the experiments. Our method was evaluated across various N-way K-shot segmentation scenarios, utilizing mean Intersection-over-Union (mIoU) as the performance metric, specifically calculated on the test classes. In alignment with the MFNET approach, we excluded the background class from the mIoU calculation and relegated foreground misclassification to be under false positives or false negatives.

\subsection{Dataset}
We based our experiments on the COCO-$20^i$ dataset, a recognized benchmark for few-shot semantic segmentation (FSS) derived from the MS COCO dataset. We assessed our model on multi-class few-shot segmentation tasks ranging from 1-way up to 20-way. The COCO-$20^i$ dataset divides its 80 classes into 4 non-overlapping folds, each containing 20 classes. To ensure a fair comparison with other methodologies, we evaluated SAC solely on the validation images for each fold, deliberately omitting the training images that other methods might have used for training. We conducted tests with 1000 randomly selected episodes for each fold and setting, aligning our performance evaluation and experimental setup with other benchmarks to facilitate direct comparisons.

\subsection{Ablation Study}
\input{tables/oneway}
We conducted an ablation study on fold-0 of the COCO-$20^i$ dataset using 1-way 1-shot and 5-way 1-shot configurations to evaluate each component's contribution to our method's overall performance. We assessed the impact of utilizing different backbones (DINOv2 versus SAM) as feature extractors during the construction of Class-Representative Feature Arrays (CRFAs). Subsequently, we systematically eliminated various components of our method—specifically, background region proposal (BGRP), intra-class filtering, and inter-class filtering—to observe their individual effects on performance. 

\section{Results and Discussion}

\subsection{Reevaluating the need for gradient learning in task adaptation vis-\'a-vis automated prompting}

We contend that foundation models can be adapted for segmentation tasks without gradient learning by employing alternative strategies that could match or exceed the performance of gradient-learning approaches. SegGPT \cite{Wang2023SegGPT:Context} showed that foundation models could be adapted to different tasks by configuring the prompt input but requires a specialized prompt encoder trained with gradient-learning. Our work demonstrates that for multi-class few-shot segmentation tasks, prompting a state-of-the-art vision foundation model is sufficient and outperforms gradient-learning approaches. Our method builds upon Matching Networks \cite{Vinyals2016MatchingLearning} which pioneered the idea of using cosine similarity to match feature maps of the query image with features from a small support set of support images containing the desired class instance. This affinity-learning is generalizable because the model can be applied to new class instances that were not available during the model development phase (through selection of the support images) unlike prototype-learning where the model is tuned specifically to class instances that were available only at gradient-learning time. PerSAM \cite{Zhang2023PersonalizeShot} evolved this concept to a prompt-only methodology free from gradient-learning. PerSAM generates a confidence map from matching the features of the query with the support set and derives a single positive and negative point prompt for Segment Anything Model (SAM) \cite{Kirillov2023SegmentAnything}. The confidence map is further embedded into input tokens of the SAM mask decoder to modulate decoder cross-attention and generate the segmentation mask.

Our method SAC extends PerSAM to generate multiclass segmentation. Our innovation is the algorithm that converts the confidence map to multiple points prompts where each point can prompt for a different class instance (in contrast to PerSAM which generates 1 pair of positive and negative point prompts for only a single class). We find that good multiclass segmentation can be achieved with only prompt generation and does not require simultaneous feature embedding to the mask-decoder cross-attention layer. Instead, the foundation model that generates the feature map of the query and support images is of critical importance. We benchmark our method with leading multiclass FSS models namely DCAMA and Label Anything (LA); firstly on the binary FSS problem and subsequently on multiclass FSS on the COCO-20i dataset.

\input{figures/qualitative}
\subsection{Prompt engineering performs competitively against gradient-learning in binary few-shot segmentation}

% Restricted datasets and leveraging pre-trained good feature representations.
% K-shot learning as the way forward
% Affinity learning rather than prototype learning. we are doing prototype learning.
% Meta-learning or in-context learning.
% \textcolor{red}{K-shot learning on pre-trained network. Not learning the specific task but learning generalizations. Previously called meta-learning (generalizable weights) and now in-context learning. In the specific case of segmentation, called referred segmentation whereby the network is not trained to recognize specific labels (prototype learning) but learns to identify features of a new label from k-shot examples (affinity learning).}

In Table \ref{tab:oneway}, we present the performance of our method, SAC, on the binary few-shot segmentation (FSS) task across 1-shot and 5-shot scenarios. Our method demonstrates competitive performance, closely matching SegGPT (which achieved 56.1 mIoU) by within 10\% mIoU, and DCAMA by within 2\%mIoU, while surpassing LA. Notably, SegGPT, DCAMA, and LA rely on gradient-based learning for the mask generator and employ pre-trained feature representation networks. Both SegGPT and LabelAnything also utilize prompts via a trainable prompt-encoder. It is worth highlighting that state-of-the-art binary FSS models \cite{Wang2023SegGPT:Context,Sun2024VRP-SAM:Prompt}, such as SegGPT and VRP-SAM, leverage pre-trained feature representation networks to achieve superior performance over specialist models like DCAMA, which uses ResNet pre-trained on ImageNet, LA utilizing VIT-B/16 pre-trained on ImageNet21k, VRP-SAM leveraging the Segment Anything Model (SAM), and SegGPT employing VIT-L. This trend suggests the potential of large foundational models to further enhance segmentation performance, as evidenced by the results from SegGPT.

Non-promptable methods like DCAMA are adapted to the specific set of classes in the training dataset; expanding the model to include new classes requires retraining the model with gradient-learning and rewriting the network weights carries the risk of compromising the performance of prior-learnt classes (catastrophic forgetting). Conversely, promptable models like SegGPT and LA offer the flexibility to introduce new classes through prompt updates, without necessitating model retraining. This ensures the preservation of prior class performance by reverting to previous prompts since model weights remain untouched. This capability facilitates the seamless integration of new classes, maintaining both model availability and historical performance.

Despite these advancements, state-of-art promptable models still require specialized prompt-encoders that must be instantiated or adapted using gradient-learning with a training dataset. In contrast, our SAC method fully eschews gradient learning. The question we seek to investigate in this work is the extent to which we can repurpose prompt-encoders from a foundation model without any kind of gradient learning to replace or adapt it. Our SAC method demonstrates that the gradient-learning free method is gradually emerging as a competitive alternative to SegGPT and VRP-SAM as it significantly outperforms PerSAM.

\input{tables/allway}
\subsection{Prompt engineering outperforms gradient-learning in multi-class few shot segmentation}

In multiclass (N-way) inference, neural networks must generate discriminative features for distinct classes based on the small set of support images. We observe in state-of-the-art methods that the per-class segmentation accuracy significantly degrades as the number of classes (N) grows. When weights are updated, there is a risk of displacing known information or good feature representations of one class in favor of another class. Presumably, this is due to limitations in the subset of the information contained in the training batch considered during individual gradient updates  - the very same advantages that accelerate learning with mini-batch or stochastic gradient descent. 

From Table \ref{tab:allway}, mIoU scores of both DCAMA and LA are approximately halved in 5-way FSS compared to 1-way FSS. Our SAC method significantly surpasses the performance of DCAMA and LA and achieves a widening performance gap as the number of classes grows. SAC appears to mitigate the trend of declining segmentation accuracy with increasing number of classes and is indicative of the robustness of prompt-only, gradient-learning free method. We believe that since no model weights are updated, effective feature representation cannot be displaced thus providing some immunity against catastrophic forgetting.

\subsection{Towards task adaptation using automated prompts s\'{a}ns gradient learning}
Our ablation study highlights the impact of different components in our SAC model. Table \ref{tab:configuration} shows that the feature representation network has the largest impact on segmentation accuracy. SAC uses DINOv2 to generate object features and substituting with the SAM image encoder results in a very significant performance decline. Inter-class filtering plays a critical role in removing incorrect class prompts and its elimination accounts for the next largest decline in performance. Our Region Class Assignment is still an inherently noisy algorithm thus the removal of erroneous prompts through inter-class filtering significantly improves segmentation mask accuracy. Intra-class filtering and background region proposal have more modest effects but remain vital components for maximizing overall performance. Intra-class filtering removes regions with low similarity scores while background region proposal pairs negative prompts with positive prompts resulting in enhancements to mask precision.

\input{tables/configuration}

Together with Matcher \cite{Liu2023Matcher:Matching}, our method, SAC, demonstrates the feasibility of automated prompt-only tuning of foundation models (without gradient learning) for task adaptation. Here we consider SAM as a foundation model that performs semantic segmentation of objects but in a class-agnostic manner. Matcher established the feasibility of segmenting a single, selectable type of object by adapting SAM to the binary FSS task. Our SAC method demonstrates how to segment multiple selectable object types by adapting SAM to the multiclass FSS task. Both Matcher and our SAC utilize automated algorithms that process the support image set to generate prompts for adapting the task of the foundation model. We further argue that multi-class FSS is an essential, intermediate capability for adapting SAM for panoptic segmentation \cite{Kirillov2018PanopticSegmentation}. Indeed, the ability to classify pixels as belonging to different object types is a precondition for identifying specific object instances (instance segmentation).

\section{Conclusion}
In this paper, we present Segment-Any-Class (SAC), a training-free method to automate prompt generation for Segment Anything Model (SAM) that task adapts it to perform multi-class few-shot segmentation. Our state-of-art results demonstrate that prompting-only approach may surpass the accuracy of gradient-learning methods particularly when the number of classes grows and also shows a significantly gradual decline in accuracy versus number of classes. This evidence suggests that prompt-only methods may benefit from new distinguishing and discriminative features from added classes without displacing the information from prior classes. The accuracy of our prompt-only, training-free SAC method is greatly dependent on the sophistication of the vision foundation model for image feature extraction. The SAM foundation model is sensitive to errors in the prompt class-region proposals thus necessitating inter- and intra-class filtering. Prompt only and training-free paradigm is an attractive and alternative approach to rapidly adapt foundation models for different tasks, particularly when the supporting training dataset is small. This paradigm renders the foundation models intrinsically immune to loss of features or context during task adaptation and allows for online and rapid task adaptation of foundation models.

%\input{figures/qualitative_domain}

% or


% \section{}
% % use section* for acknowledgment
% \section*{Acknowledgment}


\ifCLASSOPTIONcaptionsoff
  \newpage
\fi



%\bibliographystyle{IEEEtran}


% \end{thebibliography}
{\small
\bibliographystyle{IEEEtran}

\bibliography{references,references2}
}

%\appendices
%\section{Supplementary Tables}
%\input{tables/kway}


% \end{IEEEbiography}


%\newpage



%\vfill


\end{document}

%% file: figures/sac.tex
\begin{figure*}
  \centering
   \includegraphics[width=1.0\linewidth]{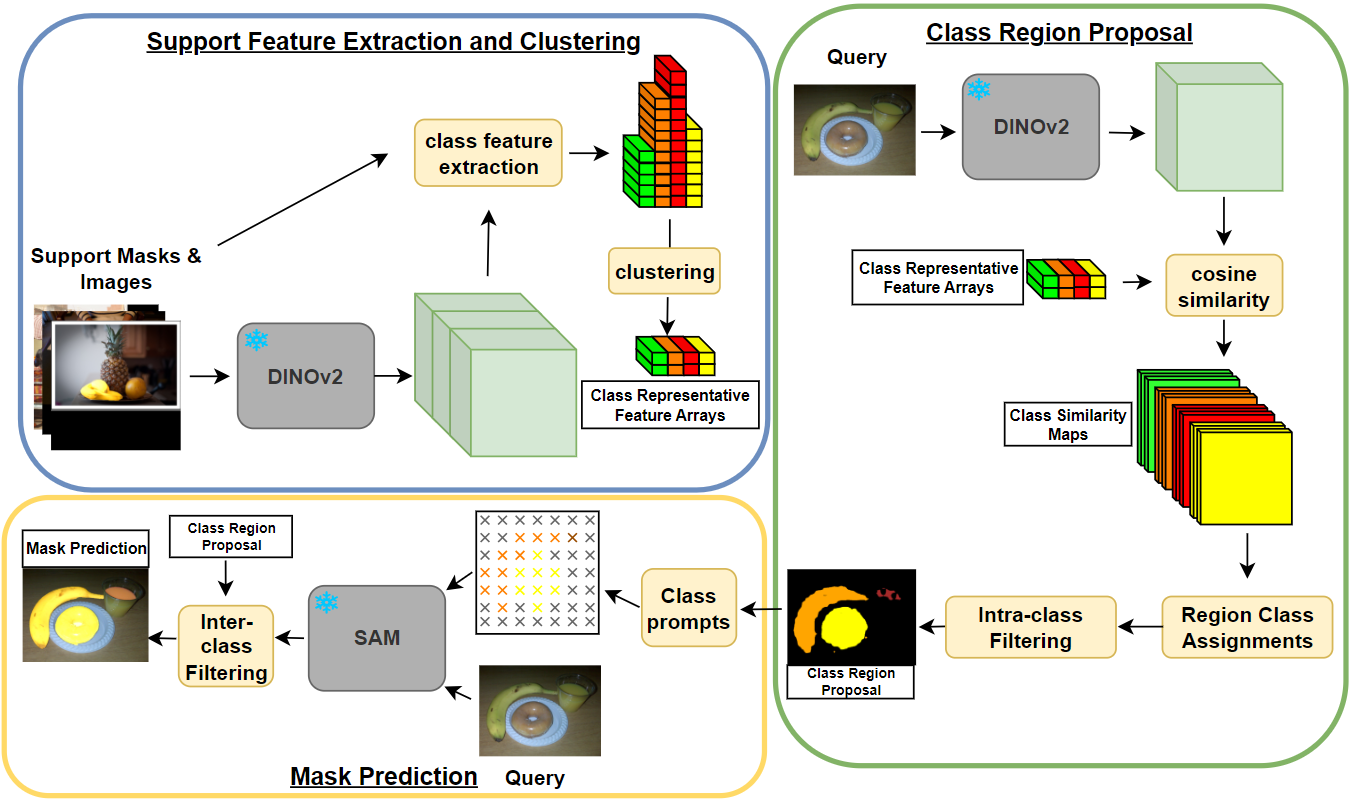}
   \caption{An overview of Segment Any-Class}
   \label{fig:sac}
\end{figure*}

%% file: tables/oneway.tex
% \usepackage{multirow}
\begin{table*}[!ht]
\centering
\caption{SAC uses automated prompt generation without modifying neural network weights and achieves competitive mIoU performance against leading gradient learning methods in 1-way few shot segmentation (FSS) on COCO-$20^i$ dataset. 
}
%1-way segmentation for Results for 1-way K-shot FSS on COCO-$20^i$. LA and DCAMA represents leading FSS methodologies and requires specific FSS training while our method is training-free.
\scalebox{1.2}{
\begin{tabular}{cccccc|ccccc}
\hline
\multirow{2}{*}{Method} & \multicolumn{5}{c|}{1-way 1-shot} & \multicolumn{5}{c}{1-way 5-shot} \\ \cline{2-11} 
 &
  \multicolumn{1}{c|}{fold-0} &
  \multicolumn{1}{c|}{fold-1} &
  \multicolumn{1}{c|}{fold-2} &
  \multicolumn{1}{c|}{fold-3} &
  mean(mIoU) &
  \multicolumn{1}{c|}{fold-0} &
  \multicolumn{1}{c|}{fold-1} &
  \multicolumn{1}{c|}{fold-2} &
  \multicolumn{1}{c|}{fold-3} &
  mean (mIoU) \\ \hline
\multicolumn{1}{c|}{DCAMA} &
  \multicolumn{1}{c|}{49.5} &
  \multicolumn{1}{c|}{52.7} &
  \multicolumn{1}{c|}{52.8} &
  \multicolumn{1}{c|}{48.7} &
  50.9 &
  \multicolumn{1}{c|}{55.4} &
  \multicolumn{1}{c|}{60.3} &
  \multicolumn{1}{c|}{59.9} &
  \multicolumn{1}{c|}{57.5} &
  58.3 \\
\multicolumn{1}{c|}{LA} &
  \multicolumn{1}{c|}{39.2} &
  \multicolumn{1}{c|}{46.6} &
  \multicolumn{1}{c|}{44.1} &
  \multicolumn{1}{c|}{42.7} &
  43.1 &
  \multicolumn{1}{c|}{42.4} &
  \multicolumn{1}{c|}{49.3} &
  \multicolumn{1}{c|}{45.9} &
  \multicolumn{1}{c|}{43.4} &
  45.1 \\
\multicolumn{1}{c|}{SAC (ours)} &
  \multicolumn{1}{c|}{45.5} &
  \multicolumn{1}{c|}{51.1} &
  \multicolumn{1}{c|}{50.9} &
  \multicolumn{1}{c|}{49.4} &
  49.2 &
  \multicolumn{1}{c|}{54.1} &
  \multicolumn{1}{c|}{58.2} &
  \multicolumn{1}{c|}{60.8} &
  \multicolumn{1}{c|}{58.9} &
  58.0
\end{tabular}
}

\label{tab:oneway}
\end{table*}

%% file: figures/qualitative.tex
\begin{figure*}
  \centering
   \includegraphics[width=0.75\linewidth]{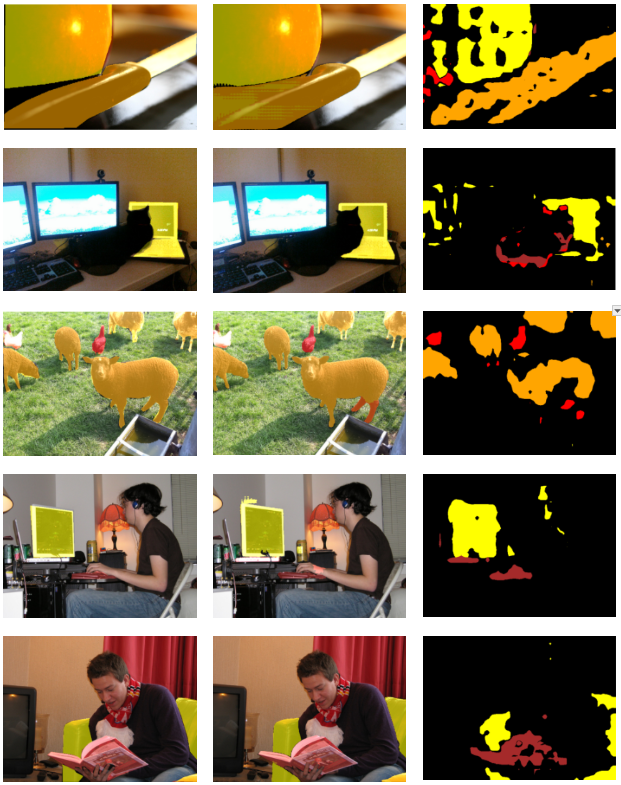}
   \caption{Qualitative results of 5-way 1-shot segmentation on COCO-$20^i$. Displayed from left to right are the following: Ground Truth, SAC prediction, and Class Region Proposal.}
   \label{fig:qualitative}
\end{figure*}

%% file: tables/allway.tex
\begin{table*}[]
\centering
\caption{Automated prompt generation with SAC outperforms mIoU of gradient learning methods in multi-way FSS on COCO-$20^i$ dataset}

%Average mIoU results of 4 folds for N-way 1-shot on COCO-$20^i$
\scalebox{1.4}{
\begin{tabular}{c|c|c|c|c|c|c|c}
\hline
Method & 1-way & 2-way & 3-way & 4-way & 5-way & 10-way & 20-way \\ \hline
DCAMA  & \textbf{50.9}  & 31.7  & 24.2  & 21.1  & 16.7  & 10.8   & 4.7    \\
LA     & 43.1  & 34.6  & 31.7  & 29.6  & 27.7  & 23.6   & 13.7   \\
SAC (ours)    & 49.2  & \textbf{46.6}  & \textbf{46.7}  & \textbf{45.0}  & \textbf{44.0}  & \textbf{41.9}   & \textbf{37.9}     
\end{tabular}
}

\label{tab:allway}
\end{table*}

%% file: tables/configuration.tex
% \begin{table}[H]
% \centering
% \caption{Ablation analysis evaluated on COCO-$20^i$ dataset demonstrate that background region proposal, intra- and inter-class filtering are critical components to SAC}
% \scalebox{1.4}{
% \begin{tabular}{c|c}
% Configuration             &     mIoU  \\ \hline
% Complete SAC               & 45.51 \\
% w/o DINOv2 (replace with SAM)                  & 26.60 \\
% w/o backgd reg prop                  & 42.00 \\
% w/o intra-class filtering & 42.83 \\
% w/o inter-class filtering & 38.93
% \end{tabular}
% }
% \label{tab:configuration}
% \end{table}

\begin{table}[H]
\centering
\caption{Ablation analysis evaluated on COCO-$20^i$ dataset demonstrate that background region proposal, intra- and inter-class filtering are critical components to SAC}
\scalebox{1.3}{
\begin{tabular}{c|cl}
Configuration                 & 1-way   & 5-way \\ \hline
w/o DINOv2 (replace with SAM) & 26.60 & 11.74     \\
w/o inter-class filtering     & 38.93 & 30.94   \\
w/o backgd reg prop           & 42.00 & 38.55 \\
w/o intra-class filtering     & 42.83 & 38.88 \\
complete SAC                  & 45.51 & 39.25 

\end{tabular}
}
\label{tab:configuration}
\end{table}